%% file: sigconf.tex
\documentclass[sigconf, nonacm]{acmart}
\AtBeginDocument{%
  }

\setcopyright{acmlicensed}
\copyrightyear{2018}
\acmYear{2018}
\acmDOI{XXXXXXX.XXXXXXX}
\acmConference[Conference acronym 'XX]{Make sure to enter the correct
  conference title from your rights confirmation email}{June 03--05,
  2018}{Woodstock, NY}
\acmISBN{978-1-4503-XXXX-X/2018/06}

\input{macros.tex}

\begin{document}

\title{Reinforcement Learning for Controller Parameters in the Auto-Bidding System at Pinterest}


\input{content/authors}
\input{content/abstract}

\ccsdesc[500]{Information systems~Online advertising}
\ccsdesc[500]{Computing methodologies~Reinforcement learning}
\ccsdesc[300]{Information systems~Auction algorithms}

\keywords{auto-bidding, reinforcement learning, PID control, budget pacing, computational advertising}

\maketitle

\input{content/introduction}

\input{content/related_work}
\input{content/problem_formulation_2}
\input{content/proposed_methods_2}
\input{content/production}

\input{content/experiment_v2}

\input{content/discussion}
\input{content/conclusion}

\begin{acks}
We would like to express our sincere gratitude to all those who contributed to the development of our RLBP model and pipeline. We especially thank Akanksha Baid, Adam Fer, and Amol Goyal for their general support \rev{in} delivering the project; Aniket Ketkar, Sameer Bhide, Jorge Alberto Bayona Romo, and Jen-An Lien for providing guidance on \rev{infrastructure} implementation; Ludek Cigler and Christina Huang for sharing their experience \rev{with} real-time \rev{bidding}. Their expertise, dedication, and collaboration were essential to the success of this
project. 
\end{acks}

\bibliographystyle{ACM-Reference-Format}
\bibliography{references}

\appendix

\end{document}

%% file: macros.tex
\usepackage{mathtools}
\usepackage{graphicx}
\usepackage{tikz}                      
\usetikzlibrary{positioning, arrows.meta, calc, backgrounds, fit}  
\usepackage{bbm}
\usepackage{amsthm}
\usepackage{xpatch}
\usepackage{xspace}
\usepackage{mathrsfs}

\usepackage{url}
\usepackage{array}
\usepackage{wrapfig}
\usepackage{multirow}
\usepackage{tabularx}
\usepackage[normalem]{ulem} 
\usepackage{enumerate}
\usepackage{enumitem}

\newif\ifpaperrevisions
\paperrevisionsfalse
\newcommand{\rev}[1]{\ifmmode\else\leavevmode\fi{\ifpaperrevisions\color{blue}\fi#1}}

\newcommand\cut[1]{{}}  
\makeatletter
\xpatchcmd{\algorithmic}{\itemsep\z@}{\itemsep=.5ex plus.5pt}{}{}
\makeatother
\usepackage{mathtools}
\mathtoolsset{showonlyrefs}

\newcommand{\cA}{{\mathcal{A}}}

\newcommand{\cS}{{\mathcal{S}}}

\newcommand{\RR}{\mathbb{R}}

\renewcommand{\iota}{I}

\newcommand{\bc}{\begin{center}}
\newcommand{\ec}{\end{center}}

\newcommand{\bdm}{\begin{displaymath}}
\newcommand{\edm}{\end{displaymath}}

\newcommand{\beq}{\begin{equation}}
\newcommand{\eeq}{\end{equation}}

\newcommand{\bfl}{\begin{flushleft}}
\newcommand{\efl}{\end{flushleft}}

\newcommand{\bt}{\begin{tabbing}}
\newcommand{\et}{\end{tabbing}}

\newcommand{\beqn}{\begin{align}}
\newcommand{\eeqn}{\end{align}}

\newcommand{\beqs}{\begin{align*}} 
\newcommand{\eeqs}{\end{align*}}  

\providecommand{\keywords}[1]
{
  \small	
  \textbf{\textit{Keywords---}} #1
}

\numberwithin{theorem}{section}

\numberwithin{assumption}{section}

\numberwithin{lemma}{section}

\makeatletter
\renewcommand*\env@matrix[1][\arraystretch]{%
  \edef\arraystretch{#1}%
  \hskip -\arraycolsep
  \let\@ifnextchar\new@ifnextchar
  \array{*\c@MaxMatrixCols c}}
\makeatother

\newenvironment{itemize*}%
{\begin{itemize}[leftmargin=*,topsep=5pt]%
		\setlength{\itemsep}{1pt}%
		\setlength{\parskip}{1pt}}%
	{\end{itemize}}

\newenvironment{enumerate*}%
{\begin{enumerate}[leftmargin=20pt,topsep=5pt]%
		\setlength{\itemsep}{1pt}%
		\setlength{\parskip}{1pt}}%
	{\end{enumerate}}

\newcommand{\agent}{\textsc{RLBP}}
\usepackage{cleveref}
\crefname{section}{\S}{\S\S}
\Crefname{section}{\S}{\S\S}

%% file: content/authors.tex
\author{Yichu Zhou}
\authornote{Authors contributed equally to this research.}
\affiliation{
\institution{Pinterest Inc.}
\city{Seattle}
\country{USA}
}
\email{yichuzhou@pinterest.com}

\author{Qinyi Chen}
\authornotemark[1]
\affiliation{
\institution{Pinterest Inc.}
\city{San Francisco}
\country{USA}
}
\email{qinyichen@pinterest.com}

\author{Fei Feng}
\authornotemark[1]
\affiliation{
\institution{Pinterest Inc.}
\city{New York}
\country{USA}
}
\email{ffeng@pinterest.com}

\author{Jacob Loh}
\affiliation{
\institution{Pinterest Inc.}
\city{San Francisco}
\country{USA}
}
\email{jloh@pinterest.com}

\author{Mehdi Ben Ayed}
\affiliation{%
\institution{Pinterest Inc.}
\city{New York}
\country{USA}
} 
\email{mbenayed@pinterest.com}

\author{Lin Yang}
\affiliation{%
\institution{Pinterest Inc.}
\institution{University of California, Los Angeles}
\city{Los Angeles}
\country{USA}
}
\email{linyang@pinterest.com}

\author{Andrei Curelea}
\affiliation{
\institution{Pinterest Inc.}
\city{San Francisco}
\country{USA}
}
\email{acurelea@pinterest.com}

\author{Jiajing Xu}
\affiliation{%
\institution{Pinterest Inc.}
\city{San Francisco}
\country{USA}
}
\email{jiajing@pinterest.com}

\renewcommand{\shortauthors}{Zhou et al.}

%% file: content/abstract.tex
\begin{abstract}
Auto-bidding systems in online advertising must balance competing objectives—maximizing budget delivery while minimizing cost per action—across heterogeneous campaigns under non-stationary market conditions. This trade-off evolves over campaign lifespans: early on, cost optimization dominates when budgets roll over; near deadlines, budget delivery becomes urgent. While PID controllers are widely used in production for their stability and interpretability, fixed tunings cannot adapt to diverse campaigns and shifting traffic patterns. We present \agent \xspace (Reinforcement Learning for Bidding Parameters), a hybrid architecture deployed at Pinterest that uses RL as a meta-controller to select campaign-specific PID parameters daily while retaining PID for high-frequency bid adjustments. 
\agent \xspace trains a multi-head $Q$-network using a relative-lift reward formulation normalized against the production baseline, and performs inference-time fusion that adapts the cost–delivery trade-off to campaign states without retraining.
Through online A/B testing at Pinterest, RLBP achieves a statistically significant reduction in effective cost-per-click (eCPC) with no revenue degradation, and two months of post-rollout measurements against a holdout also show consistent cost reductions. These results demonstrate that meta-controller RL can successfully augment production control systems at scale.

\end{abstract}

%% file: content/introduction.tex
\section{Introduction}\label{sec:intro}

Pinterest is a visual discovery platform that serves approximately $619$ million monthly active users and generated \$4.2 billion in advertising revenue in 2025~\cite{pinterest2025q4}. Like every large advertising platform, its ads delivery system must solve an \emph{auto-bidding} problem on behalf of each advertiser: given a budget, a lifetime schedule, and cost-efficiency targets such as cost-per-click (CPC) or return-on-spend (ROS), decide how much to bid for each user-level impression so that the campaign delivers its budget smoothly while satisfying its KPI constraints~\cite{aggarwal2024autobidding,balseiro2024field}. Because these decisions are taken billions of times per day, even fractional improvements to the bidding policy translate into substantial advertiser value and platform revenue.

A long line of work formulates auto-bidding as a budget- and KPI-constrained optimization problem, whose optimal solution, under standard assumptions and primal--dual arguments, reduces to bidding a scalar multiple of the predicted impression value~\cite{wu2018budget,he2021uscb,yang2019multivariable,balseiro2024field,aggarwal2024autobidding,chen2024ebaypacing}. The operational question is then how to update that scalar \emph{online}, under non-stationary traffic and in the face of daily variation in user behavior, advertiser entry, and competitive pressure. In production, this online update is \rev{often implemented using} a proportional--integral--derivative (PID) feedback controller that tracks the running error between realized performance and a pacing target and adjusts a bid multiplier to close the gap~\cite{zhang2016feedback,agarwal2014linkedin,yang2019multivariable,stram2024mystique,karlsson2020frontier}. PID controllers are simple, interpretable, and \rev{widely used}, which is why they remain the backbone of industrial pacing systems at LinkedIn, Yahoo, Alibaba, and elsewhere. Their weakness, however, is rigidity: a single fixed setup cannot be simultaneously appropriate for hundreds of thousands of heterogeneous campaigns and shifting market conditions.

Reinforcement learning (RL) is a natural way to replace such rules with a policy that adapts to campaign states. A large body of work formulates real-time bidding as a Markov decision process and trains a deep RL agent to emit bids directly~\cite{cai2017drlb,wu2018budget,jin2018dcmab,he2021uscb,wen2022maab,guo2024aigb}. In production, however, a black-box policy that outputs bids \rev{using} advertisers' budgets faces two stubborn obstacles. First, \emph{safety and explainability}: neural policies can fail catastrophically (severe under- or over-delivery, cost performance degradation), and such failures are hard to diagnose and remediate immediately and can lead to unrecoverable revenue impact and terrible advertiser experience~\cite{korenkevych2024offline}. Second, \emph{off-policy learning}: \rev{because} of safety \rev{concerns}, online exploration with random bids is often unacceptable, so policies are constrained to be trained from logged data, exposing the agent to the full difficulty of offline RL~\cite{kumar2020cql}.

We present \agent~(\emph{Reinforcement Learning for Bidding Parameters}), the auto-bidding pacing system deployed and evaluated with live traffic at Pinterest. The central design choice is to \emph{not} use RL to produce bids directly. Instead, we retain the existing PID controller and use RL as a \emph{meta-controller} that configures the controller. Specifically, at the beginning of each day and for each campaign, the RL agent observes the campaign state, produces a set of controller parameter multipliers and materializes these decisions as a lightweight lookup table for serving. This table is fixed for the next 24 hours. During online serving, the PID controller consumes the same table and continues to perform fine-grained bid adjustment at the serving cadence. We deliberately restrict the RL agent to operate on a small, bounded set of controller multipliers. This confines its decisions to modest adjustments around a validated production configuration, rather than allowing it to generate bids directly or select unconstrained controller settings. Beyond safety, this design is also lightweight in infrastructure cost, since only a flat lookup table---not a model artifact---is loaded at serving time. Related designs include \citet{korenkevych2024offline}, who use offline RL to optimize the parameters of an existing base policy at Meta, and \citet{yang2025drlput}, who apply deep RL to tune the ranking utility function at Pinterest. \textsc{RLBP} shares the philosophy of augmenting an existing production bidder with a learned layer.

Our main contributions are listed below:
\begin{itemize}[leftmargin=*,topsep=2pt,itemsep=1pt]
    \item \textbf{A hybrid RL--PID architecture for production auto-bidding.} We learn campaign-specific PID parameter multipliers while retaining high-frequency feedback control. No model inference at serving time is required.
    \item \textbf{\rev{Multi-objective balancing without retraining.}} We train a multi-head $Q$-network for separate goals, budget utilization and cost efficiency, and then fuse the values at inference time for policy induction. This design adapts the delivery--cost trade-off dynamically based on campaign states.
    \item \textbf{Relative-lift rewards for clearer learning guidance.} Instead of using absolute values of the objective metrics (e.g. revenue or cost in dollars), we define rewards as the metric lift against the production baseline. This transformation not only normalizes different metrics to a comparable scale but also reshapes the reward distribution and provides a clearer learning signal for the agent to improve policy performance. 
    \item \textbf{End-to-end production deployment and empirical validation at Pinterest.} We design an exploration--training--serving pipeline that integrates seamlessly into Pinterest's production bidding system with negligible infrastructure overhead. The online A/B test shows a statistically significant $0.25\%$ eCPC reduction with no revenue degradation. The post-rollout measurements against holdout traffic demonstrate consistent cost reduction and click-through rate improvement.
\end{itemize}

The remainder of the paper is organized as follows. Section~\ref{sec:related_work} reviews related work. Section~\ref{sec:formulation} describes the formulation of the meta-controller MDP. Section~\ref{sec:proposed_methods} details the central techniques including the multi-head DQN, reward shaping, and adaptive fusion. Section~\ref{sec:production} covers production integration, and Section~\ref{sec:exp} presents offline and online experiments, followed by concluding remarks.

%% file: content/related_work.tex
\section{Related Work}\label{sec:related_work}

\subsection{Auto-Bidding and Budget Pacing}
Auto-bidding is the problem of bidding in real-time auctions on behalf of advertisers to maximize campaign value under budget and KPI constraints~\cite{aggarwal2024autobidding,balseiro2024field,lucier2024autobidders}. Classical analyses reduce the optimal bid to a scalar multiple of the predicted impression value, so the operational task reduces to online estimation of that scalar~\cite{wu2018budget,he2021uscb,yang2019multivariable}. Production systems realize this in two ways. The \emph{feedback-control} lineage drives the running spend error to a pacing target via online feedback---from probabilistic throttling on a participation probability at LinkedIn~\cite{agarwal2014linkedin}, to PI-style multiplicative bid control in Yahoo's \emph{Mystique}~\cite{stram2024mystique}, Alibaba's multivariable controller~\cite{yang2019multivariable}, and the RTB work of~\citet{zhang2016feedback}; \citet{karlsson2020frontier} surveys this line. The \emph{optimization} lineage instead solves a parametric bid-shading program online via dual-based algorithms, such as those of Balseiro et al.~\cite{balseiro2024field} and Chen et al.~\cite{chen2024ebaypacing}. \citet{xu2015smart} take a complementary approach that allocates delivery rate across traffic segments. \rev{Our work complements these pacing mechanisms by learning campaign-specific controller settings and adapting the delivery--cost trade-off over} campaign lifecycles.

\subsection{Reinforcement Learning for Bidding}
Starting with~\citet{cai2017drlb}'s MDP formulation of real-time bidding, a substantial line of work has applied deep RL to produce bids directly. \citet{wu2018budget} introduce reward shaping tailored for budget-constrained bidding; \citet{jin2018dcmab} and~\citet{wen2022maab} address the multi-agent setting; \citet{he2021uscb} unify constrained bidding under a DDPG-based controller; \citet{guo2024aigb} reformulate bidding as generative trajectory modeling via diffusion. On the offline-RL side, \citet{kumar2020cql} provide the canonical CQL estimator, \citet{mou2022sustainable} propose an online RL framework with a safety-constrained exploration policy that bridges the offline--online gap, and \citet{korenkevych2024offline} deploy at Meta a hybrid agent whose neural network is trained offline to adjust an existing base policy and then discarded at serving time. In the Pinterest setting specifically, \citet{yang2025drlput} apply deep RL to tune the \emph{ranking utility} function, a different subsystem from pacing. Across most of this line, the learned policy ultimately determines the bid itself, which creates deployment friction in safety-sensitive production environments.

\subsection{Hybrid RL and Classical Control}
Outside advertising, combining RL with classical controllers has a long history in robotics and process control: a line of work uses RL to \emph{tune} the gains of a PID controller rather than replace it, trading off adaptivity against stability~\cite{lawrence2022stabilitypid,mdpipid2025,grinding2025}; the broader safe-RL and constrained-MDP literature studies explicit safety constraints~\cite{altman1999cmdp,garcia2015survey}. The closest work to ours is \citet{korenkevych2024offline}, who augment Meta's production bidder with offline RL. Our focus is daily, campaign-specific selection over a discrete set of PID gain multipliers around the production configuration. We further implement a multi-head $Q$-network for separate business goals and use adaptive fusion to combine the values at inference time.

Overall, our work sits at the intersection of the three lines above: we retain the PID feedback-control structure, borrow offline training and IPS-based screening~\cite{horvitz1952ipw,wang2017optimal} from the RL lineage, and validate the design through A/B tests on live Pinterest traffic.

%% file: content/problem_formulation_2.tex
\section{Problem Formulation}\label{sec:formulation}
This section formalizes the auto-bidding problem as a reinforcement learning task. We first describe how PID controllers are used in production bidding systems~(\Cref{sec:pid_controller}), then present our meta-controller MDP formulation where the RL agent learns to select PID controller parameters~(\Cref{sec:mdp_pid}).

\subsection{PID Controller for Auto Bidding}\label{sec:pid_controller}
Using PID controllers for online bidding is prevalent in industrial applications~\cite{agarwal2014linkedin,zhang2016feedback,yang2019multivariable,stram2024mystique}. Based on \rev{campaign objectives}, the controller can be designed to fulfill the budget smoothly or achieve desired cost performance. The core idea is to track the error between the current performance metric (e.g., spend rate, cost efficiency) and a target value, then adjust the bid accordingly to minimize this error over time.

Once the target is given, at time step $\ell$, the controller adjusts the next bid value $b_{\ell+1}$ with
\begin{equation}
b_{\ell+1} = f(b_\ell, ~k_p \cdot errP_\ell, ~k_i \cdot errI_\ell, ~k_d \cdot errD_\ell),
\label{eq:pid_update}
\end{equation}
where the function $f$ encapsulates the update formula, which typically follows a conventional summation or multiplicative approach. The terms $errP_{\ell}, errI_{\ell}, errD_{\ell}$ represent the proportional, integral, and derivative error terms, respectively. The parameters $k_p, k_i, k_d \in \RR_+$ are known as the controller \emph{coefficients}, which determine the relative influence of each error term on the bid adjustment.
\cut{
\begin{itemize}[leftmargin=*,topsep=2pt,itemsep=1pt]
\item \textbf{Proportional term} $errP_t$ captures the current error between the actual metric and the target;
\item \textbf{Integral term} $errI_t = \sum_{\tau=0}^t errP_\tau$ accumulates historical errors to eliminate steady-state bias;
\item \textbf{Derivative term} $errD_t$ measures the rate of change of the error to dampen oscillations.
\end{itemize}
}

Although the PID controller has been adopted for many decades, it remains a practical first-choice option for bidding systems due to its stability, interpretability, and effectiveness. However, a critical challenge is dynamically selecting appropriate parameters: a fixed set of parameters cannot simultaneously accommodate the diverse characteristics of different campaigns (e.g., budget size, lifetime, traffic patterns) or adapt to dynamic market conditions.

\subsection{MDP for Controller-based Bidding}\label{sec:mdp_pid}

We adopt the standard Markov decision process (MDP) formulation throughout the paper. Specifically, an MDP is defined by a state space $\cS$, a finite action space $\cA$, \rev{a transition kernel $P(s^{\prime}\mid s,a)$,} a reward function $r: \cS \times \cA \rightarrow \mathbb{R}$, and a discount factor $\gamma \in [0,1)$. A policy $\pi(a \mid s)$ specifies the probability of taking action $a \in \cA$ at state $s \in \cS$, and induces the action-value function
\begin{equation}
Q^{\pi}(s,a) := \mathbb{E}^{\pi}\!\left[\sum_{t=0}^{\infty} \gamma^t r(s_t,a_t)\,\middle|\, s_0=s,\, a_0=a\right].
\end{equation}

The objective is to learn a policy that maximizes the expected cumulative discounted reward, or equivalently, the corresponding \rev{$Q$}-values over the state-action space.

We formulate the online bidding process as an MDP, where the key design is to use RL as a \emph{meta-controller} which only selects PID parameters $(k_p, k_i, k_d)$ rather than directly producing bids. 
In our implementation, the agent takes one step per day. We use $t$ to index daily RL decisions and $\ell$ to index the controller updates in Eq.~\eqref{eq:pid_update}. At the start of each day, the agent observes each campaign's state and selects the corresponding parameters $(k_p,k_i,k_d)$. The PID controller then uses these parameters to update the campaign's bids according to Eq.~\eqref{eq:pid_update} throughout that day. Practitioners can use a different inference cadence as needed.

\textbf{State Space.} The state $s_t \in \cS$ encodes the campaign's status at the beginning of day $t$, including the campaign \rev{ID} embedding, lifetime remaining days, daily budget, PID controller targets (e.g., a campaign's target spend curve), and previous-day aggregated metrics (CTR, CPC, revenue).
These features collectively provide the agent with information about the campaign's lifecycle stage, performance history, and operational constraints, enabling it to make informed decisions about appropriate controller tuning.

\textbf{Action Space.} Each action $a_t = (m_p, m_i, m_d) \in \cA$ specifies a multiplicative adjustment to the default controller parameters $(k_p^0, k_i^0, k_d^0)$. Concretely, the deployed parameters are given by
\begin{equation}
(k_p, k_i, k_d) = (m_p \cdot k_p^0,\; m_i \cdot k_i^0,\; m_d \cdot k_d^0).
\label{eq:gain_multipliers}
\end{equation}


In our implementation, we optimize only the proportional ($m_p$) and integral multipliers ($m_i$), while keeping the derivative multiplier ($m_d$) fixed at 1. This design choice is motivated by the fact that the derivative error signal $errD_{\ell}$ is fairly noisy, making derivative control less reliable for policy optimization. 
For each of the two tuned dimensions, the multiplier is chosen from the discrete set $\rev{\{}1-\alpha,\, 1,\, 1+\alpha\rev{\}}$, where $\alpha \in (0,1)$ controls the maximum deviation from the production baseline. For example, $\alpha=0.2$ corresponds to allowing $\pm 20\%$ adjustments around the default setting. The resulting action space therefore contains $3^2=9$ actions. Both the number of actions and the allowable deviation are intentionally \emph{kept small} to limit online exploration to bounded changes around the validated configuration.

\textbf{Transition Dynamics.} After the agent selects action $a_t$ at the start of day $t$, the PID controller uses the adjusted parameters from Eq.~\eqref{eq:gain_multipliers} to serve bids throughout the day. The next state $s_{t+1}$ is determined by the aggregated outcomes of these bids: total spend, clicks, revenue, and remaining budget. The transition $P(s_{t+1}|s_t, a_t)$ is stochastic due to the inherent randomness in user behavior, auction competition, and ad delivery dynamics. Importantly, the environment is \emph{non-stationary}: market conditions, user engagement patterns, and competitor strategies evolve daily, making adaptive policies crucial for sustained performance.

\textbf{Reward Function.} The reward $r(s_t, a_t)$ quantifies the campaign's performance over day $t$ under the selected PID parameters. We aim to balance two primary objectives:
\begin{enumerate}[leftmargin=*,topsep=2pt,itemsep=1pt]
\item \textbf{Maximizing budget utilization (BU)}: \rev{Spending} the \rev{campaign's} allocated budget to maximize revenue delivery;
\item \textbf{Minimizing cost per click (CPC)}: Achieving cost-efficient delivery to meet advertiser ROI expectations.
\end{enumerate}
Note that these two objectives can easily conflict with each other. Higher bids can increase delivery at the expense of higher CPC, while lower bids may improve cost efficiency but leave part of the budget
unspent. 
The intricacy here is to balance the two objectives based on campaign states, which we will detail later.
\cut{Rather than directly using absolute metric values (e.g., utilization rate $\approx 0.95$, CPC $\approx \$0.50$), which yield weak learning signals due to small inter-action differences, we adopt a \emph{relative lift} reward formulation that compares each action's performance against a production baseline. This reward shaping mechanism, detailed in Section~\ref{sec:reward_reshape}, significantly improves policy convergence by amplifying the distinction between good and bad actions.}

\textbf{Discount Factor.} 
In our deployment, we instantiate this RL framework with $\gamma = 0$, yielding a one-step contextual-bandit objective based on daily outcomes. This choice avoids bootstrapping errors in offline training, and the cross-day trade-off between cost efficiency and delivery is instead handled by the lifecycle-dependent fusion weight (see \Cref{sec:adaptive_reward}). It is a deployment choice rather than a restriction of the meta-controller architecture: practitioners can use $\gamma \in (0, 1)$ and train the $Q$-heads to estimate discounted returns for longer-term optimization, while retaining the state, action, and controller interface. We discuss this extension in \Cref{sec:discussion}.




%% file: content/proposed_methods_2.tex
\section{Proposed Methods}
\label{sec:proposed_methods}
In this section, we present \agent, an RL framework for ads bidding that jointly optimizes budget utilization and cost efficiency. We first describe the multi-head $Q$-networks (\Cref{sec:two_heads_arch}), then explain how its predictions are fused (\Cref{sec:adaptive_reward}) and converted into a bidding-control policy (\Cref{sec:policy_induction}). Finally, we specify the relative-lift rewards used to train the agent (\Cref{sec:reward_reshape}).

\subsection{Multi-head $Q$-Networks}
\label{sec:two_heads_arch}
Since we have multiple objectives for the agent to optimize, we define separate reward functions, e.g., reward for budget utilization and reward for cost reduction. A conventional approach is to combine these rewards into a scalar and train a $Q$-network to predict the resulting value.
However, in production ads bidding systems, the relative importance of different optimization objectives often changes over the lifetime of a campaign. Early in a campaign, since unspent budget can roll over to future days, we can prioritize cost performance. Near the final day of a campaign, spending the remaining budget becomes a more immediate concern. As a result, the desired trade-off between cost efficiency and revenue generation is inherently dynamic rather than fixed. Such a requirement of flexibility in objective weighting can be troublesome under a conventional RL paradigm where the reward is a fixed function. To obtain different policies, practitioners often need to redefine the reward and retrain the model, which is computationally expensive and operationally inconvenient in a production environment. 

To address this issue, we propose a multi-head $Q$-network. The model feeds a common state representation into multiple separate branches, each predicting $Q$-values for one objective. As illustrated in  \Cref{fig:dqn_arch}, we have a head for budget utilization that outputs $Q^{bu}(s,a)$ and a head for CPC that outputs $Q^{cpc}(s,a)$. The outputs are fused only at inference to derive a policy. Depending on how the predicted $Q$-values are fused, we can generate different policies from a single trained model. 


\input{content/figures/model_arch}

\subsubsection{Loss Functions}
The total loss comprises temporal difference losses from all heads. In our case, given a training dataset $\{(s_i,a_i,r^{bu}_i, r^{cpc}_i, s'_i)\}_{i=1}^N$, the per-objective losses and the total loss respectively are
\begin{align}
& L^{bu}(\theta) = \frac{1}{N}\sum_{i=1}^N
\left(Q_{\theta}^{bu}(s_i, a_i)-r_i^{bu}-\gamma\cdot\max_{a\in\cA}Q_{\theta}^{bu}(s'_i, a)\right)^2 \\
& L^{cpc}(\theta) = \frac{1}{N}\sum_{i=1}^N\left(Q_{\theta}^{cpc}(s_i, a_i)-r_i^{cpc}-\gamma\cdot\min_{a\in\cA}Q_{\theta}^{cpc}(s'_i, a)\right)^2 \\
& L^{total}(\theta) = ~L^{bu}(\theta) + L^{cpc}(\theta)
\end{align}
where $\theta$ denotes the model parameters; $r_i^{bu}$ and $r_i^{cpc}$ are the budget utilization and CPC rewards; $N$ is the total number of training examples. 
The CPC head uses $\min$ in its Bellman backup because lower CPC lift indicates better cost performance.
In our deployment, with $\gamma=0$, each $Q$-head is trained to fit the corresponding daily rewards.

\subsubsection{State Representation}
The state representation is constructed from the state definition in \Cref{sec:formulation}. Before being fed into the $Q$-network, the \textit{campaign's target spend curve} is normalized into an hourly distribution, and budget features are log-transformed. Categorical features are embedded, while all vector and scalar numerical features are concatenated and batch-normalized. The final state representation is the concatenation of the categorical embeddings and normalized numerical feature block.

\subsection{Inference-time Adaptive $Q$-Value Fusion}\label{sec:adaptive_reward}

We define the fused $Q$-value as a fusion of the two heads:
\begin{equation}
\hat{Q}(s,a) = Q^{bu}(s,a) \mathbin{-} \beta(d) \cdot Q^{cpc}(s,a) \label{eq:Q_fuse}
\end{equation}
where $d$ denotes the number of remaining days in the campaign's lifetime,
and $\beta(d) \geq 0$ is a cost-penalty weight that balances budget
utilization against cost reduction. The negative sign reflects that we
seek to maximize budget utilization while minimizing CPC. We design
$\beta(d)$ to decrease as $d$ approaches zero, progressively reducing
the cost penalty and shifting the policy's emphasis toward budget
utilization as the campaign approaches its deadline. We compare
alternative adaptive weighting functions in \Cref{sec:exp_ablation}.

This inference-time adaptive fusion design provides two main advantages. First, it separates outcome prediction from business preferences by learning each objective separately, avoiding the difficulty of encoding a changing trade-off in a single reward function. Second, it offers practical flexibility in deployment, since the trade-off can be adjusted directly through the fusion weight $\beta{(d)}$ as campaign conditions evolve. This makes \agent~better suited for real-world ad systems, where optimization priorities are often non-stationary and must be updated in a lightweight and interpretable manner. 

\subsection{Policy Induction}\label{sec:policy_induction}

Given the fused scores in \Cref{sec:adaptive_reward}, we construct a policy over the discrete joint action space.

Rather than using a greedy policy that always selects the highest-scoring action, we adopt a top-$K$ stochastic policy. Specifically, the policy retains the $K$ actions with the highest fused scores, forming the state-dependent candidate set $A^{\prime}(s)$. A temperature-scaled softmax is applied over this reduced action set:
\begin{equation}
    \pi(a\vert s) = \frac{\exp\left(\hat{Q}(s,a)/\tau\right)}{\sum\limits_{a^{\prime}\in A^{\prime}(s)}\exp\left(\hat{Q}(s,a^{\prime})/\tau\right)} \qquad \text{for~} a \in A'(s).
\end{equation}
where $\hat{Q}(s,a)$ is the fused score after numerical preprocessing and $\tau > 0$ is the temperature; the probability is zero outside $A^{\prime}(s)$. The final bidding-control decision is then sampled from this distribution.

\subsection{Reward Shaping}\label{sec:reward_reshape}

We now specify the daily rewards used to train the multi-head $Q$-networks in \Cref{sec:two_heads_arch}.

Learning directly from raw business metrics, especially CPC, presents significant challenges. Since the CPC difference between candidate actions is often small relative to its absolute value (for example, \$1.01 versus \$1.00 per click),  training the $Q$-network to accurately rank these actions can be difficult. Furthermore, CPC ranges vary widely across campaigns, and campaigns with larger absolute prediction errors can disproportionately dominate the loss function during training.

An additional challenge stems from the varying scales across distinct reward signals. For instance, CPC values may span from 0 to 100, whereas budget utilization remains strictly bounded within the interval $\left[0, 1\right]$. Such scale discrepancies generate imbalanced gradient signals backpropagated to the shared network layers during training, while simultaneously introducing complexity when tuning fusion weights at inference.


\input{content/figures/reward_gap}

For policy selection, the key question is how candidate actions compare under the two objectives. We therefore shape the reward to the metric lift over the production baseline rather than the raw metric value. Formally, let $v(s,a)$ denote the observed metric (e.g., CPC) under state $s$ and action $a$, and let $v^{prod}(s)$ denote the corresponding metric estimated from matched observations under the production baseline. CPC is total spend divided by total clicks, whereas BU is total spend divided by allocated budget. Instead of using $v(s,a)$ directly as the target reward, we define the reward as:
\begin{equation}
    r(s,a) = \frac{v(s,a)- v^{prod}(s)}{v^{prod}(s)}
\end{equation}
Applied to BU and CPC, this definition yields the two training rewards $r^{bu}$ and $r^{cpc}$ used in \Cref{sec:two_heads_arch}, so that each $Q$-function learns the relative metric change of an action over the baseline system. 
Their outputs are then combined through the fusion rule to construct
the action-selection policy..

This reward shaping provides two advantages. First, it centers targets around the existing controller and expresses deviations on a relative scale. Under squared-error training, normalization by the reference $v^{prod}(s)$ weights squared prediction errors by $1/(v^{prod}(s))^2$, emphasizing relative discrepancies across campaigns. \Cref{fig:reward_gap} illustrates the relative action-gap distributions under the two reward formulations. In our offline study, we also evaluate their effect on policy outcomes; see \Cref{tab:ablation_reward}. Second, the lift values of different metrics, such as CPC and BU, tend to lie on more comparable scales than their raw absolute values. This makes it easier to balance multiple objectives in the inference-time $Q$-value fusion discussed in \Cref{sec:adaptive_reward}, since the combination weights no longer need to compensate for large differences in metric magnitude.

Overall, by shaping the reward as metric lift, \agent~learns
baseline-relative action values rather than absolute metric values.
This design aligns target scales across campaigns and simplifies multi-objective trade-off control at inference time.

%% file: content/figures/model_arch.tex
\definecolor{onlineblue}{RGB}{102,102,255}
\definecolor{onlinefill}{RGB}{246,246,255}

\definecolor{offlineorange}{RGB}{255,153,72}
\definecolor{offlinefill}{RGB}{255,247,238}

\definecolor{servinggreen}{RGB}{0,128,0}
\definecolor{servingfill}{RGB}{244,255,244}

\definecolor{boxfill}{RGB}{250,239,226}
\definecolor{linegray}{RGB}{25,25,25}

\tikzset{
  panelblue/.style={
    draw=onlineblue,
    fill=onlinefill,
    rounded corners=7pt,
    line width=0.75pt
  },
  block/.style={
    draw=linegray,
    fill=boxfill,
    rounded corners=4pt,
    line width=0.75pt,
    align=center,
    minimum width=2.65cm,
    minimum height=0.78cm,
    inner sep=3pt,
    font=\rmfamily\large
  },
  stateblock/.style={
    draw=offlineorange,
    fill=offlinefill,
    rounded corners=4pt,
    line width=0.85pt,
    align=center,
    minimum width=6.9cm,
    minimum height=0.88cm,
    inner sep=3pt,
    font=\rmfamily\large
  },
  fusionblock/.style={
    draw=servinggreen,
    fill=servingfill,
    rounded corners=4pt,
    line width=0.85pt,
    align=center,
    minimum width=4.2cm,
    minimum height=0.82cm,
    inner sep=3pt,
    font=\rmfamily\large
  },
  arrow/.style={
    -{Stealth[length=6pt,width=6pt]},
    line width=0.75pt,
    draw=black
  },
  headblock/.style={block},
  inferencepanel/.style={draw=servinggreen, fill=servingfill, rounded corners=4pt},
  dashedarrow/.style={arrow, dashed},
  plainline/.style={
    line width=0.75pt,
    draw=black
  }
}

\begin{figure}
\centering

\resizebox{0.95\linewidth}{!}{%
\begin{tikzpicture}[font=\rmfamily]

\draw[panelblue]
  (-5.55,1.08) rectangle (5.55,6.55);

\node[
  fill=white,
  text=black,
  font=\rmfamily\bfseries\normalsize,
  inner sep=2pt
] at (-5.55,3.82)
  {Repeat $\times 3$};


\node[stateblock] (state) at (0,0)
  {State Representation};

\node[block] (l1) at (-2.45,1.75) {Linear layer};
\node[block] (l2) at (-2.45,2.78) {Layer Norm};
\node[block] (l3) at (-2.45,3.81) {Leaky ReLU};
\node[block] (l4) at (-2.45,4.84) {Dropout};
\node[block] (l5) at (-2.45,5.87) {Linear Layer};

\node[block] (r1) at (2.45,1.75) {Linear layer};
\node[block] (r2) at (2.45,2.78) {Layer Norm};
\node[block] (r3) at (2.45,3.81) {Leaky ReLU};
\node[block] (r4) at (2.45,4.84) {Dropout};
\node[block] (r5) at (2.45,5.87) {Linear Layer};

\node[headblock] (qbu) at (-2.45,7.15) {$Q^{\mathrm{bu}}$ Head};
\node[headblock] (qcpc) at (2.45,7.15) {$Q^{\mathrm{cpc}}$ Head};

\draw[inferencepanel]
  (-2.7,8.45) rectangle (2.7,9.95);

\node[
  fill=white,
  text=servinggreen,
  font=\rmfamily\bfseries\normalsize,
  inner sep=2pt
] at (0,9.95)
  {Inference only};

\node[fusionblock] (fusion) at (0,9.15)
  {Adaptive Fusion};


\coordinate (split) at (0,0.96);

\draw[plainline] (state.north) -- (split);
\draw[arrow] (split) -| (l1.south);
\draw[arrow] (split) -| (r1.south);

\draw[arrow] (l1.north) -- (l2.south);
\draw[arrow] (l2.north) -- (l3.south);
\draw[arrow] (l3.north) -- (l4.south);
\draw[arrow] (l4.north) -- (l5.south);
\draw[arrow] (l5.north) -- (qbu.south);

\draw[arrow] (r1.north) -- (r2.south);
\draw[arrow] (r2.north) -- (r3.south);
\draw[arrow] (r3.north) -- (r4.south);
\draw[arrow] (r4.north) -- (r5.south);
\draw[arrow] (r5.north) -- (qcpc.south);

\draw[dashedarrow]
  (qbu.north) -- ++(0,0.35)
  -| ([xshift=-0.65cm]fusion.south);

\draw[dashedarrow]
  (qcpc.north) -- ++(0,0.35)
  -| ([xshift=0.65cm]fusion.south);

\end{tikzpicture}
}

\caption{\rev{Multi-head $Q$-network architecture.}}
\label{fig:dqn_arch}
\end{figure}
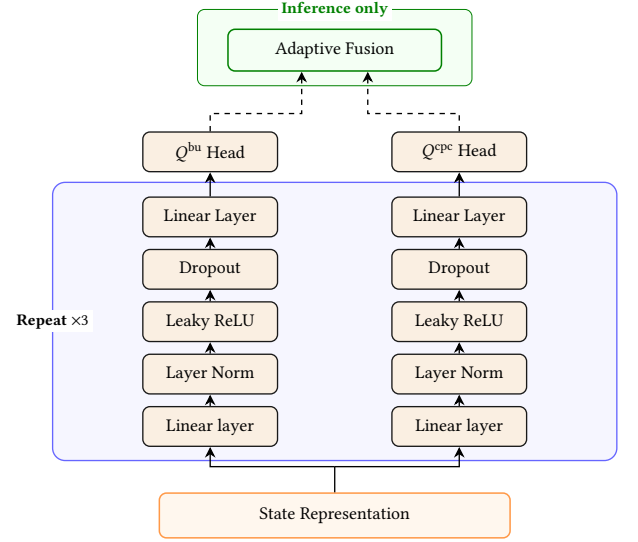

%% file: content/figures/reward_gap.tex
\begin{figure}[] 
    \centering
    \includegraphics[width=1.0\linewidth]{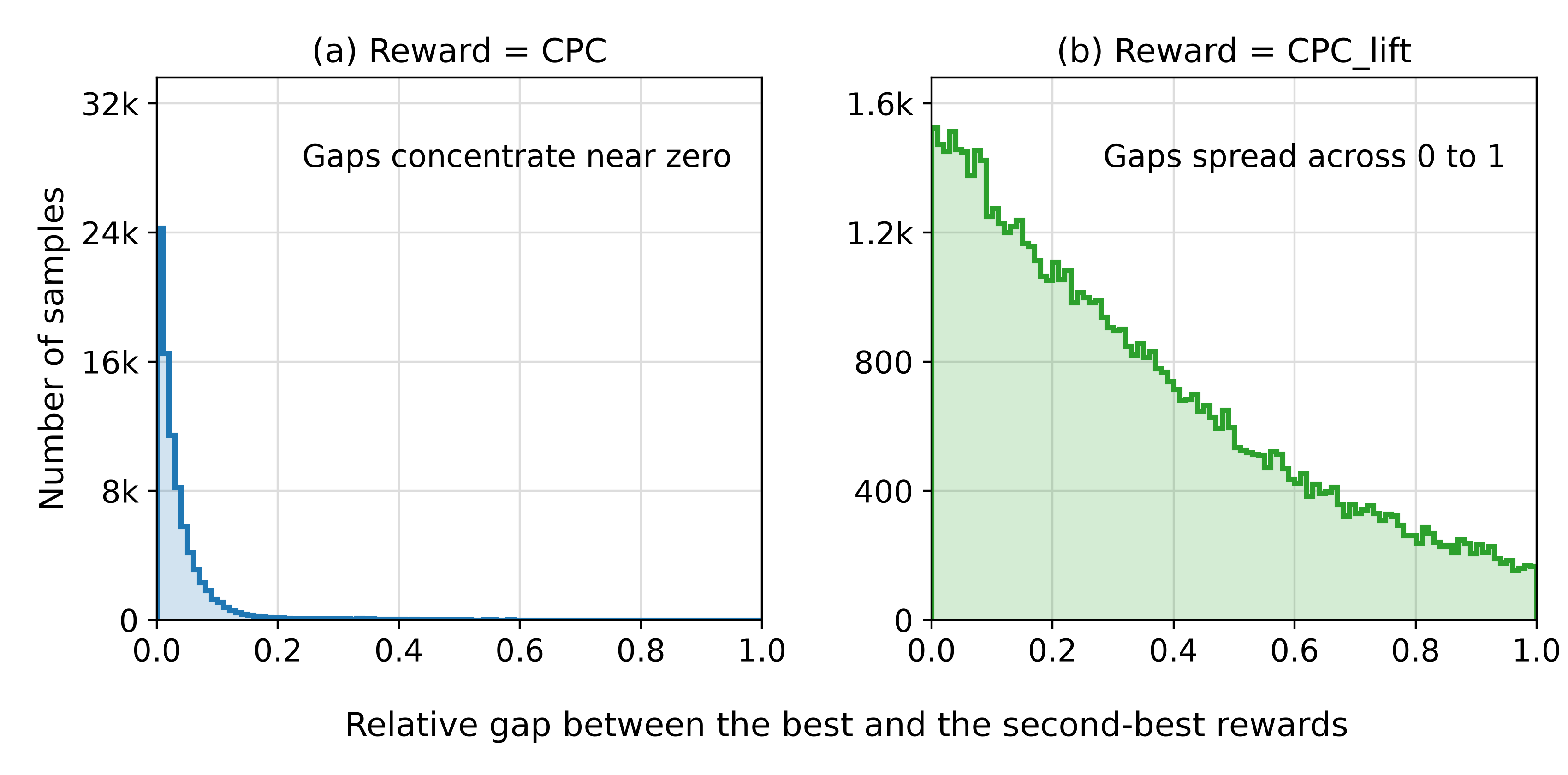}
    \caption{Reward relative gap distribution comparison between absolute metric values vs. metric lifts. The left panel shows the relative difference between the best and second-best rewards constructed with raw CPC, while the right panel shows the corresponding relative difference by using CPC lift. }
    \label{fig:reward_gap}
\Description{Two histograms show the relative gap between the best and second-best rewards. Absolute CPC gaps concentrate near zero; CPC lift gaps span more of the interval.}
\end{figure}

%% file: content/production.tex
\section{Production Integration}\label{sec:production}
This section describes the integration of \agent~into Pinterest's production auction and bidding system. As illustrated in \Cref{fig:pipeline}, the pipeline comprises three stages: online exploration-and-exploitation, offline training, and daily serving.

\input{content/figures/pipeline}

\paragraph{Online exploration-and-exploitation.}
Training the DQN requires \rev{logged} tuples $(s_t, a_t, r_t^{bu}, r_t^{cpc}, s_{t+1})$ collected under a known logging policy. 
While serving the learned policy, we reserve a dedicated portion of campaign traffic and
budget for uniform random exploration. Specifically, each campaign's budget is split into two segments: the primary portion is allocated to the RL policy for exploitation, while the remaining portion is used for the uniform random exploration, with assignments refreshed daily. Each segment uses distinct user buckets and separate budgets and bids. This separation allows exploratory actions to be evaluated within their allocated traffic and budget. 



\paragraph{Offline training.}
A batch training job periodically aggregates the transition samples over a historical time window and retrains the $Q$-network. 
The model learns separate $Q$-functions for the target objectives using the reward formulation described in Section \ref{sec:proposed_methods}. 
Once training is complete, the resulting checkpoint is used to generate serving-time decisions for active campaigns.

\paragraph{Daily serving.}
A scheduled job runs \agent~once per day over all eligible campaigns: it computes $Q$\rev{-}values for every $a \in \cA$, fuses the two heads by the adaptive rule in \Cref{sec:adaptive_reward}, selects an action under the top-\rev{$K$} stochastic policy of \Cref{sec:policy_induction}, and materializes the chosen multipliers as a campaign-level lookup table (\Cref{fig:rl-pid-arch}). The production bidder then adjusts bids on the order of seconds, fetching the corresponding controller parameters from that table and applying them through the existing controller; it never loads the DQN or invokes the agent at request time, which minimizes inference latency.

\input{content/figures/meta-controller}

%% file: content/figures/pipeline.tex
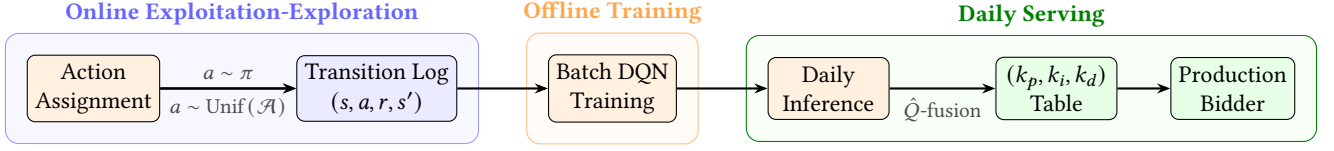
\begin{figure*}[htbp]
\centering
\begin{tikzpicture}[
    node distance=0.6cm and 0.7cm,
    box/.style={rectangle, draw, rounded corners=3pt, minimum height=0.8cm, minimum width=1.6cm, align=center, font=\normalsize},
    data/.style={box, fill=blue!8},
    proc/.style={box, fill=orange!12},
    serve/.style={box, fill=green!10},
    arrow/.style={-{Stealth[length=5pt]}, thick},
    phase/.style={font=\normalsize\bfseries},
    lbl/.style={font=\small, text=black!70},
]

\node[proc] (hash) {Action\\Assignment};
\node[data, right=1.8cm of hash] (log) {Transition Log\\$(s,a,r,s')$};

\node[proc, right=1.2cm of log] (train) {Batch DQN\\Training};

\node[proc, right=1.2cm of train] (infer) {Daily\\Inference};
\node[serve, right=1.4cm of infer] (table) {$(k_p,k_i,k_d)$\\Table};
\node[serve, right=of table] (lcb) {Production\\Bidder};

\draw[arrow] (hash) -- node[below, lbl] {$a \sim \mathrm{Unif}(\cA)$} (log);
\draw[arrow] (hash) -- node[above, lbl] {$a \sim \pi$} (log);
\draw[arrow] (log) -- (train);
\draw[arrow] (train) -- (infer);
\draw[arrow] (infer) -- node[below, lbl] {$\hat{Q}$-fusion} (table);
\draw[arrow] (table) -- (lcb);


\begin{scope}[on background layer]
    \node[draw=blue!40, fill=blue!3, rounded corners=5pt, inner sep=8pt,
          fit=(hash)(log), label={[phase, blue!60]above:Online Exploitation-Exploration}] {};
    \node[draw=orange!50, fill=orange!3, rounded corners=5pt, inner sep=8pt,
          fit=(train), label={[phase, orange!70]above:Offline Training}] {};
    \node[draw=green!50!black, fill=green!3, rounded corners=5pt, inner sep=8pt,
          fit=(infer)(table)(lcb), label={[phase, green!50!black]above:Daily Serving}] {};
\end{scope}

\end{tikzpicture}
\caption{End-to-end training-serving pipeline of \agent. 
}
\label{fig:pipeline}
\end{figure*}

%% file: content/figures/meta-controller.tex
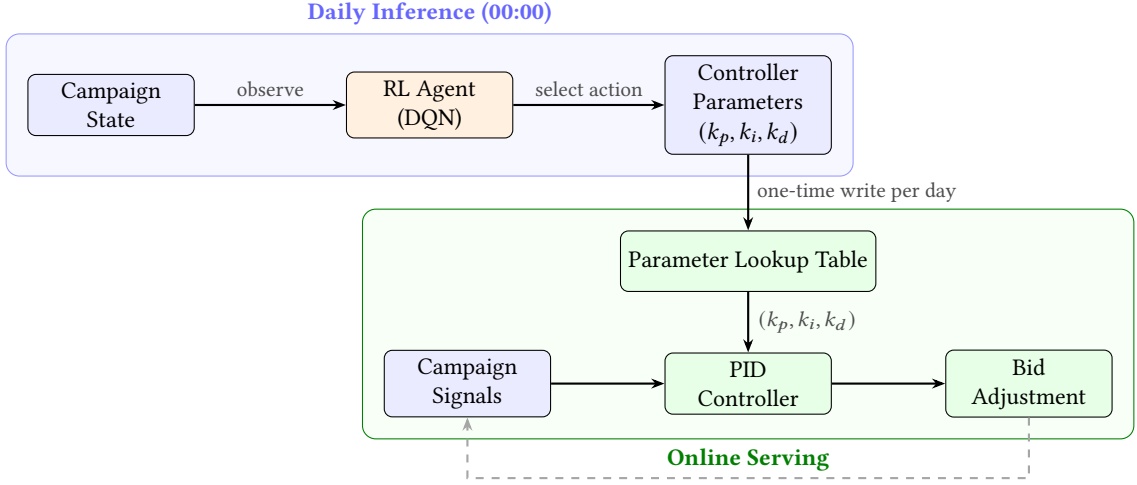
\begin{figure*}[htbp]
\centering
\begin{tikzpicture}[
    node distance=0.6cm and 0.7cm,
    box/.style={rectangle, draw, rounded corners=3pt, minimum height=0.8cm, minimum width=2.2cm, align=center, font=\normalsize},
    data/.style={box, fill=blue!8},
    proc/.style={box, fill=orange!12},
    serve/.style={box, fill=green!10},
    arrow/.style={-{Stealth[length=5pt]}, thick},
    phase/.style={font=\normalsize\bfseries},
    lbl/.style={font=\small, text=black!70},
]

\node[data] (state) {Campaign\\State};
\node[proc, right=2.0cm of state] (agent) {RL Agent\\(DQN)};
\node[data, right=2.0cm of agent] (gains) {Controller\\Parameters\\$(k_p,k_i,k_d)$};

\node[serve, below=1.0cm of gains] (table) {Parameter Lookup Table};
\node[serve, below=0.8cm of table] (pid) {PID\\Controller};
\node[data, left=1.5cm of pid] (request) {Campaign\\Signals};
\node[serve, right=1.5cm of pid] (bid) {Bid\\Adjustment};

\draw[arrow] (state) -- node[above, lbl] {observe} (agent);
\draw[arrow] (agent) -- node[above, lbl] {select action} (gains);
\draw[arrow] (gains) -- node[right, lbl] {one-time write per day} (table);

\draw[arrow] (request) -- (pid);
\draw[arrow] (table) -- node[right, lbl] {$(k_p,k_i,k_d)$} (pid);
\draw[arrow] (pid) -- node[above, lbl] {} (bid);

\draw[arrow, dashed, gray!70] (bid.south) -- ++(0,-0.8) -| node[below, lbl, pos=0.25] {} (request.south);

\begin{scope}[on background layer]
    \node[draw=blue!40, fill=blue!3, rounded corners=5pt, inner sep=8pt,
          fit=(state)(agent)(gains), label={[phase, blue!60]above:Daily Inference (00:00)}] {};
    \node[draw=green!50!black, fill=green!3, rounded corners=5pt, inner sep=8pt,
          fit=(request)(pid)(table)(bid), label={[phase, green!50!black]below:Online Serving}] {};
\end{scope}

\end{tikzpicture}
\caption{Two-layer RL+PID architecture. The RL agent observes campaign state once daily and selects controller gains from a small, bounded set. The PID controller uses those fixed gains for all bid adjustments over the next 24 hours, \rev{retaining the existing feedback-control structure}.}
\label{fig:rl-pid-arch}
\end{figure*}

%% file: content/experiment_v2.tex
\section{Experiments}\label{sec:exp}

In this section, we evaluate \agent~both offline and online. In the offline study, we evaluate the design choices and screen candidate configurations against the production baseline. In the online experiment, we validated the efficacy of \agent~through A/B testing on Pinterest's production advertising platform. 

\subsection{Offline Study}\label{sec:exp_offline}

\subsubsection{IPS-based offline lift}\label{sec:exp_ips}
We use 21 days of transition data for training and 7 days of \rev{held-out} data for evaluation. Let $\hat{\pi}(a\mid s)$ denote the known logging propensity. Suppose \rev{the held-out} dataset \rev{contains} $N$ examples, \rev{$\{(s_i,a_i,r_i, s'_i)\}_{i=1}^N$}, where $s_i$ is the state, $a_i$ is the action, \rev{$r_i$ is} the observed \rev{reward}, and $s'_i$ is the next state. For a policy $\pi$, we can estimate its value with IPS:
\begin{equation} 
\mathrm{IPS}(\pi) = \frac{1}{N}\sum_{i=1}^{N} \frac{\pi(a_i \mid s_i)}{\hat{\pi}(a_i\mid s_i)}\cdot r_i \label{eq:ips} 
\end{equation}
The offline results below use uniformly randomized exploration
logs, with $\hat{\pi}(a \mid s) = 1/|\mathcal{A}|$. After rollout (\Cref{sec:exp_rollout}), the logging policy mixes
the deployed policy with uniform exploration.




To measure the improvements, we compare the IPS of an offline-trained policy $\pi$ against the production policy $\pi_{\text{prod}}$, i.e.,
\begin{equation} 
\mathrm{lift}(\pi) = \frac{\mathrm{IPS}(\pi) - \mathrm{IPS}(\pi_{\mathrm{prod}})}{\mathrm{IPS}(\pi_{\mathrm{prod}})}. 
\label{eq:offline_lift}
\end{equation} 
In the following section, we present the IPS lift achieved by each technique. Because our framework incorporates dual reward functions—one designed to boost budget utilization and the other to minimize costs—we report the corresponding IPS lifts separately for each objective.


\subsubsection{Ablation study}\label{sec:exp_ablation}
We isolate the contribution of two core techniques of \agent: the relative-lift reward formulation (\Cref{sec:reward_reshape}) and the choice of adaptive function $\beta(d)$ in the inference-time \rev{$Q$}-fusion (\Cref{sec:adaptive_reward}). In each ablation, all other components are held fixed at \agent's deployed configuration and results are averaged across five training runs with different random seeds. Bold rows mark the production choice.

\input{content/tables/ablation_reward}

\paragraph{Reward Shaping.}
We compare training with absolute metric values against relative lift as the reward signal. 
\Cref{tab:ablation_reward} quantifies the final policy quality: relative lift achieves $2.77\%$ revenue lift on the final campaign day and \rev{$3.02\%$} CPC reduction on earlier days, compared to $0.08\%$ and \rev{$2.17\%$} for absolute rewards. 
The difference in revenue-lift is especially pronounced.



\input{content/tables/ablation_adaptive}

\paragraph{Adaptive function.}
We compare three forms of the fusion weight $\beta(d)$ (\Cref{fig:adaptive_shapes}a): a \emph{step} function switching abruptly at $d=1$, a \emph{linear} ramp $\beta(d) \propto d$, and a \emph{sigmoid} between them. All \rev{increase} cost emphasis \rev{with} remaining days, differing in \rev{their} transition \rev{profiles}. \Cref{fig:adaptive_shapes}b shows the consequence for policy behavior: greater cost emphasis yields more CPC-reducing actions, gradually under the linear schedule and abruptly under the step.

\input{content/figures/adaptive_function_shapes}

\Cref{tab:ablation_adaptive} reports the overall IPS lift on a cohort of campaigns in the last week of their lifetime, and \Cref{fig:adaptive_tradeoff} breaks down CPC and revenue lift by lifetime remaining days across the same cohort.
\rev{Day-level lifts vary across the lifecycle.} The linear schedule is the only \rev{evaluated} variant with \rev{both nonnegative} revenue lift \rev{($+0.05\%$) and reduced CPC ($-0.34\%$) in the point estimates} (\Cref{tab:ablation_adaptive}), so we adopt it in production. \rev{Sigmoid has higher revenue ($+1.41\%$) alongside higher CPC ($+0.49\%$).}

\input{content/figures/fusion_func}


\subsection{Online A/B Experiments}\label{sec:exp_online}

We evaluate \agent~in a seven-day live A/B test using the following deployed configuration: relative-lift reward, linear adaptive fusion, and top-\rev{$K$} softmax inference. All    
experiment arms share the same serving infrastructure; the production baseline uses the default controller with no RL-driven adjustment. 
Against this baseline, we report revenue, clicks, and effective cost-per-click (eCPC). 
We report results at two levels of aggregation: \textit{all campaigns}, which includes every treated campaign-day and reflects overall online impact, and 
\textit{lifetime-ended campaigns}, restricted to campaigns whose lifetime completed during the experiment. The latter captures campaign-end behavior using observations within the test window.

\subsubsection{Overall results}
\Cref{tab:online_overall} summarizes online performance at two levels of aggregation. Across all treated campaign-days, \agent \xspace achieves a statistically significant \rev{$0.25\%$} reduction in eCPC \rev{($p<0.05$)} while keeping revenue \rev{approximately unchanged}, indicating that the learned policy routes spend more cost-efficiently. \rev{For lifetime-ended campaigns, revenue and budget utilization are particularly relevant because the opportunity to spend the remaining budget ends with the campaign. In this cohort,} we observe a $+0.43\%$ revenue lift and $+0.25\%$ click lift\rev{, alongside $+0.18\%$ eCPC lift}. \rev{This pattern is consistent with the intended shift toward delivery near campaign end.} 


\input{content/tables/online_overall}


\subsubsection{Segmented analysis}

We further break results down by budget level and historical CPC tier (\Cref{tab:online_segmented}). eCPC \rev{point estimates} are \rev{negative} across all \rev{reported} segments, with the largest gains in the high-budget ($-0.90\%$ eCPC, $+0.81\%$ clicks) and high-CPC ($-0.64\%$ eCPC) segments where there is the most room for optimization. 

\input{content/tables/online_segmented}

\subsection{Post-Rollout Performance}\label{sec:exp_rollout}

Following the online A/B experiments in \Cref{sec:exp_online}, we measured \agent's performance over two full months after rollout at Pinterest against a small slice of \emph{holdout control} traffic using default parameters $(k_p^0, k_i^0, k_d^0)$. This extends the evaluation beyond the seven-day A/B test to a larger campaign population.

\paragraph{Measurements.}
We divide lifetime-budget click-optimized campaigns into two groups for evaluation: campaign-days with seven or more remaining days of lifetime, and those with fewer. This division aligns with our linear adaptive fusion deployment, in which the weighting progressively transitions from cost efficiency to budget utilization starting when seven days remain (\Cref{sec:adaptive_reward}). Since the holdout control has much less traffic and budget, we compare BU (instead of absolute revenue), eCPC, and click-through rate (CTR). Within each \rev{group} we sum spend, budget, clicks and impressions over the campaigns active on each day, form that day's metrics, and average across days with equal weight. Due to the imbalanced traffic, no statistical test is conducted.

\paragraph{Results.} The results for the two lifecycle groups are consistent with the linear adaptive fusion function (see \Cref{tab:online_rollout}). Where cost receives the greatest emphasis, \agent~achieves nearly $1\%$ lower eCPC, with slightly lower BU than the control; where the weighting turns toward BU, delivery is above the control while eCPC remains lower. These patterns are consistent with both earlier offline and online evaluations: offline, the linear schedule was selected because it reduced CPC while yielding nonnegative revenue lift; online, the A/B test showed an eCPC reduction overall, with a positive revenue-lift point estimate among lifetime-ended campaigns. These figures are a directional reference on long-run behavior, with \Cref{sec:exp_online} remaining the controlled estimate.

\input{content/tables/online_rollout}

%% file: content/tables/ablation_reward.tex
\begin{table}[htbp]
\centering
\caption{Reward formulation ablation. Each row reports IPS lift (\%) over the production baseline. Revenue lift is evaluated on the final day of the campaign lifetime ($d=0$), where the policy prioritizes \rev{budget} utilization; CPC lift is evaluated on $d > 0$, where cost efficiency is the primary objective. \textbf{Bold} marks the production choice.}
\label{tab:ablation_reward}
\begin{tabular}{@{}lcc@{}}
\toprule
Reward variant & Revenue lift ($d{=}0$) & CPC lift ($d{>}0$) \\
\midrule
Absolute reward       & {0.08\%} & {-2.17\%} \\
\textbf{Relative-lift reward} & {\textbf{2.77\%}} & {\textbf{-3.02\%}} \\
\bottomrule
\end{tabular}
\end{table}

%% file: content/tables/ablation_adaptive.tex
\begin{table}[htbp]
\centering
\caption{Adaptive function ablation. Each row reports IPS lift (\%) on a cohort of campaigns tracked across their full last-seven-day lifecycle. \textbf{Bold} marks the production choice.}
\label{tab:ablation_adaptive}
\begin{tabular}{@{}lcc@{}}
\toprule
Adaptive function $\beta(d)$ & Revenue lift & CPC lift \\
\midrule
Step  & $-1.11\%$ & $-1.23\%$ \\
\textbf{Linear} & $\mathbf{+0.05}\%$ & $\mathbf{-0.34}$\% \\
Sigmoid    & $+1.41\%$ & $+0.49\%$ \\
\bottomrule
\end{tabular}
\end{table}

%% file: content/figures/adaptive_function_shapes.tex
\begin{figure}[htbp]
\centering
\includegraphics[width=0.90\columnwidth]{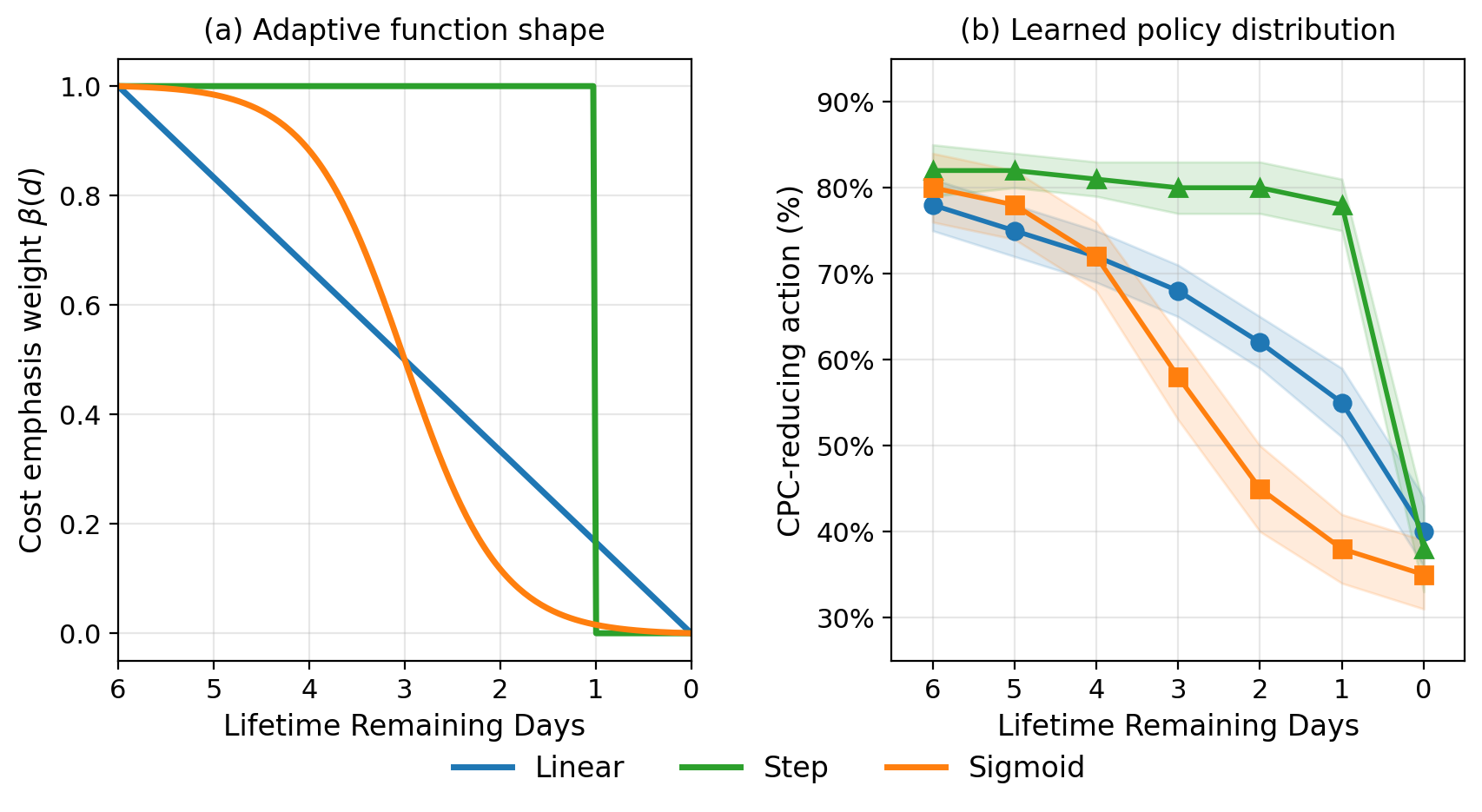}
\caption{(a) Three adaptive functions $\beta(d)$ controlling cost emphasis as a function of lifetime remaining days. (b) Resulting learned policy behavior: fraction of campaigns assigned a cost-reduction action under each adaptive function.}
\label{fig:adaptive_shapes}
\end{figure}

%% file: content/figures/fusion_func.tex
\begin{figure}[htbp]
\centering
\includegraphics[width=0.90\columnwidth]{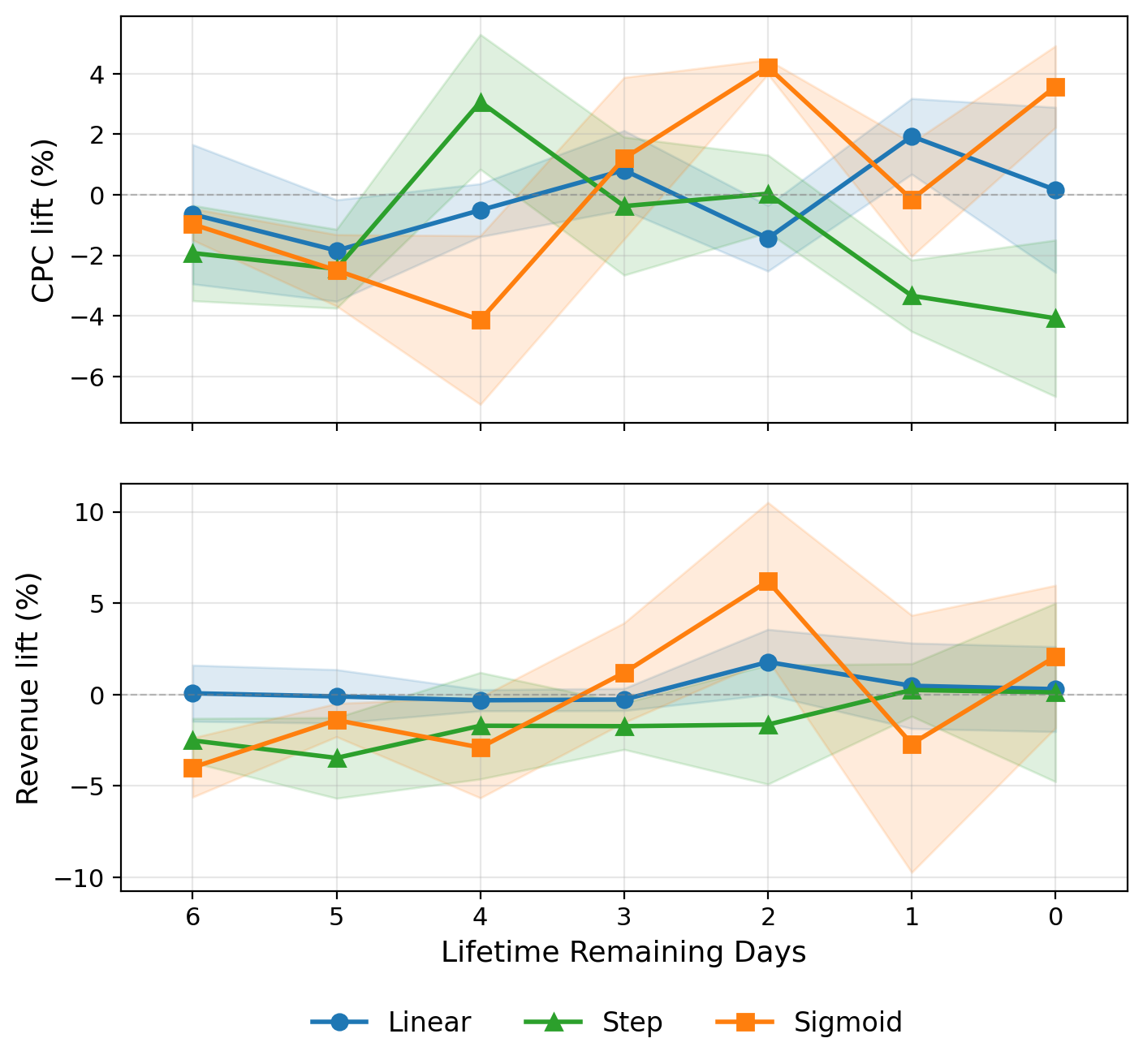}                                \caption{Offline IPS CPC and revenue lift (\%) by lifetime remaining days for different adaptive functions, evaluated on campaigns in the last seven days of their lifetime. 
}  
\label{fig:adaptive_tradeoff}
\end{figure}

%% file: content/tables/online_overall.tex
\begin{table}[htbp]
\centering
\caption{Online A/B results. Bold numbers \rev{are statistically significant ($p$-value $<0.05$).}}
\label{tab:online_overall}
\begin{tabular}{@{}lcc@{}}
\toprule
Metric & All campaigns & Lifetime-ended campaigns \\
\midrule
Revenue            & flat & +0.43\% \\
Clicks             & +0.19\% & +0.25\% \\
eCPC               & \textbf{-0.25\%} & +0.18\% \\
\bottomrule
\end{tabular}
\end{table}

%% file: content/tables/online_segmented.tex
\begin{table}[htbp]
\centering
\caption{eCPC and click lifts in online A/B experiments, segmented by budget level and historical CPC tier. Bold numbers \rev{are statistically significant ($p$-value $<0.05$).}}
\label{tab:online_segmented}
\begin{tabular}{@{}lcccc@{}}
\toprule
& \multicolumn{2}{c}{Budget level} & \multicolumn{2}{c}{Historical CPC tier} \\
\cmidrule(lr){2-3} \cmidrule(lr){4-5}
Metric & Low & High & Low & High \\
\midrule
eCPC lift   & -0.66\% & \textbf{-0.90\%} & -0.29\% & \textbf{-0.64\%} \\
Clicks lift & +0.02\% & +0.81\% & +0.12\% & +0.10\% \\
\bottomrule
\end{tabular}
\end{table}

%% file: content/tables/online_rollout.tex
\begin{table}[htbp]
\centering
\caption{Post-rollout performance of \agent~relative to the control holdout by two lifecycle stages. BU denotes budget utilization (delivered spend over budget).}
\label{tab:online_rollout}
\begin{tabular}{@{}lccc@{}}
\toprule
Lifecycle stage & BU & eCPC & CTR \\
\midrule
$\geq 7$ days remaining & -0.59\% & -0.98\% & +0.66\% \\
$< 7$ days remaining    & +0.60\% & -0.72\% & +0.00\% \\
\bottomrule
\end{tabular}
\end{table}

%% file: content/discussion.tex
\section{Discussion}\label{sec:discussion}

\cut{\rev{We discuss lessons learned from deploying \agent~at Pinterest, their implications for related research and industrial systems, and opportunities for future extensions.}

\paragraph{\rev{RL for controller parameters.}}
\rev{The key technical choice in \agent~is to use RL as a meta-controller that selects campaign-specific PID gain multipliers. Learning at the parameter level adapts the controller's response to campaign state while preserving the existing bid-execution path. Separating daily policy updates from high-frequency feedback control enables production deployment without adding model inference to the auction path. This experience suggests that learning over the controls an existing system already exposes can offer a practical route to deploying RL.}

\paragraph{\rev{Interpreting the delivery--cost trade-off.}}
\rev{Our dual-$Q$-head design separates outcome prediction from business preferences, allowing the delivery--cost trade-off to change with campaign stage without retraining. The A/B results illustrate why both objectives must be interpreted together:} among lifetime-ended campaigns\rev{,} revenue \rev{and} eCPC \rev{lifts have positive point estimates, consistent with} emphasizing delivery near campaign end. Utilization and cost efficiency compete, so a policy's value depends on the trade-off appropriate to the campaign's objective and lifecycle stage rather than on improvement in any single metric. \rev{A broader lesson is to make changing business priorities explicit in both policy design and evaluation.}}

In this section, we discuss several directions for extending the RLBP framework in industrial systems.

\paragraph{Multi-step learning.} As a first step toward industrial deployment, we instantiate RLBP with $\gamma = 0$. Nevertheless, this framework is not confined to a single-step setup. 
The meta-controller architecture, the reward design, and the multi-head $Q$-network (trained with temporal-difference errors) with adaptive inference-time fusion all extend naturally to multi-step settings, enabling long-horizon optimization. One caveat in the multi-step setting is that each head bootstraps through its own greedy policy, so the fused $Q$-value is generally not the optimal $Q$-value of the combined reward. Fusion then serves as a practical heuristic, and developing fusion schemes with formal guarantees is a promising direction for future work.

\paragraph{Expanded action space.} In this work, we limit RL actions to a small number of multipliers around the production default parameters mainly due to safety considerations. 
This narrows down the scope of exploration and simplifies deployment, at the cost of adjustment precision and range.
A potential next step is to experiment with a finer action grid or continuous gain multipliers for more precise tuning, but this would require broader exploration coverage, additional safeguards, and more robust offline evaluation. 

\paragraph{More frequent inference.} Another direction of improvement is to increase the inference frequency. Currently, inference occurs once per day as an initial implementation. We are actively exploring shorter cadences, such as hourly or minute-level updates. Elevating the inference rate enables the state representation to track intra-day fluctuations, traffic spikes, and shifting market dynamics with greater precision. This allows the RL agent to adjust decisions more dynamically, thereby reducing suboptimal performance driven by update delays. However, transitioning toward near-real-time serving renders offline lookup tables unusable, requiring online model deployment and leading to higher infrastructure costs.

%% file: content/conclusion.tex
\section{Conclusion}\label{sec:conclusion}

We presented \agent, an RL meta-controller that augments (rather than replaces) a production PID auto-bidder, providing a practical path for deploying RL in safety-sensitive advertising systems. \agent~introduces a multi-head DQN with inference-time adaptive fusion and a relative-lift reward formulation. Deployed at Pinterest, \agent ~achieves statistically significant eCPC reductions \rev{with approximately unchanged revenue} in online A/B experiments, demonstrating that RL can deliver measurable value on live advertising traffic at scale. Future directions built upon the current framework include extending to multi-step optimization for longer-lived campaigns, relaxing to continuous action spaces, and generalizing to additional campaign types and controller parameters.